\documentclass[conference]{IEEEtran}
\IEEEoverridecommandlockouts
\usepackage{amsmath,amssymb,amsfonts}
\usepackage{graphicx}
\usepackage{xcolor}

\usepackage{tabularray}
\usepackage{booktabs}
\UseTblrLibrary{booktabs}
\usepackage{caption}
\usepackage{subcaption}
\usepackage[hidelinks]{hyperref}

\usepackage[
backend=biber,
style=ieee,
doi=false,
bibencoding=utf8]{biblatex}
\AtBeginBibliography{\small}

\AtEveryBibitem{
    \clearfield{urlyear}
    \clearfield{urlmonth}
}

\def\BibTeX{{\rm B\kern-.05em{\sc i\kern-.025em b}\kern-.08em
    T\kern-.1667em\lower.7ex\hbox{E}\kern-.125emX}}
\begin{document}

\title{TRIFFID: Autonomous Robotic Aid For Increasing First Responders Efficiency \\
  \thanks{Research was possible due to the funding from the European Union’s Horizon Europe research and innovation programme under grant agreement No 101168042 (TRIFFID). \\
    \textsuperscript{1}Department of Informatics and Telematics, Harokopio University of Athens,
    E-mail: \{cani, pkoletsis, kfoteinos, varlamis, g.th.papadopoulos\}@hua.gr\\
    \textsuperscript{2}Hellenic Mediterranean University,
    E-mail: g.kefaloukos@pasiphae.eu, emarkakis@hmu.gr\\
    \textsuperscript{3}Research \& Development Department, Information Technology for Market Leadership,
    E-mail: \{l.argyriou, m.falelakis\}@itml.gr\\
    \textsuperscript{4}Institut de Robòtica i Informàtica Industrial, CSIC-UPC,
    E-mail: \{idelpino, asantamaria\}@iri.upc.edu\\
    \textsuperscript{5}NTIS Research Center, University of West Bohemia,
    E-mail: \{mcech, osevera\}@ntis.zcu.cz\\
    \textsuperscript{6}Institute of Cognitive Sciences and Technologies, National Research Council, ISTC-CNR,
    E-mail: \{alessandro.umbrico, francesca.fracasso, andrea.orlandini\}@istc.cnr.it\\
    \textsuperscript{7}Telesto Technologies Ltd.,
    E-mail: dimitris@telesto.gr\\
  }
}

\author{
  \IEEEauthorblockN{
    Jorgen Cani\textsuperscript{1},
    Panagiotis Koletsis\textsuperscript{1},
    Konstantinos Foteinos\textsuperscript{1},
    Ioannis Kefaloukos\textsuperscript{2}, \\
    Lampros Argyriou\textsuperscript{3},
    Manolis Falelakis\textsuperscript{3},
    Iván Del Pino\textsuperscript{4},
    Angel Santamaria-Navarro\textsuperscript{4},
    Martin Čech\textsuperscript{5},\\
    Ondřej Severa\textsuperscript{5},
    Alessandro Umbrico\textsuperscript{6},
    Francesca Fracasso\textsuperscript{6},
    AndreA Orlandini\textsuperscript{6},\\
    Dimitrios Drakoulis\textsuperscript{7},
    Evangelos Markakis\textsuperscript{2},
    Iraklis Varlamis\textsuperscript{1},
    Georgios Th. Papadopoulos\textsuperscript{1}\\
  }
}

\maketitle

\begin{abstract}

  The increasing complexity of natural disaster incidents demands innovative technological solutions to support first responders in their efforts. This paper introduces the TRIFFID system, a comprehensive technical framework that integrates unmanned ground and aerial vehicles with advanced artificial intelligence functionalities to enhance disaster response capabilities across wildfires, urban floods, and post-earthquake search and rescue missions. By leveraging state-of-the-art autonomous navigation, semantic perception, and human-robot interaction technologies, TRIFFID provides a sophisticated system composed of the following key components: hybrid robotic platform, centralized ground station, custom communication infrastructure, and smartphone application. The defined research and development activities demonstrate how deep neural networks, knowledge graphs, and multimodal information fusion can enable robots to autonomously navigate and analyze disaster environments, reducing personnel risks and accelerating response times. The proposed system enhances emergency response teams by providing advanced mission planning, safety monitoring, and adaptive task execution capabilities. Moreover, it ensures real-time situational awareness and operational support in complex and risky situations, facilitating rapid and precise information collection and coordinated actions.

\end{abstract}


\section{Introduction}
\label{sec:introduction}

\IEEEPARstart{P}{hysical} phenomena can lead to catastrophic events that severely impact human lives \cite{directorate-generalforeuropeancivilprotectionandhumanitarianaidoperationsechoeuropeancommissionOverviewNaturalManmade2021}. With rising urbanization and climate change, these events become even more destructive. Most catastrophe-related deaths occur in urban settings like cities, towns, and suburbs \cite{StateEuropeanCities2016}. To address this, the proposed system aims to enhance preparedness and timely information diffusion to First Responders (FRs). Recently, the incorporation of Autonomous Mobile Robots (AMRs) for supporting human efforts, minimizing personnel exposure to danger, and delivering vital crisis information to FRs promptly has shown promising results.

Modern robots are not only capable of enduring harsh situations, but they are also expendable; a luxury not affordable by human operators. However, a sufficient amount of autonomy is required for the robots to be useful to FRs in a real scenario. Additionally, a remote operation mechanism is also critical in cases where the autonomous mechanism is unable to navigate through a rough situation. Robots with this set of capabilities could shield FRs from hazards during missions, thereby speeding up response times.

According to recent initiatives and experimentation, Unmanned Ground Vehicles (UGVs) and Unmanned Aerial Vehicles (UAVs) can improve situational awareness through sensors. This raw data is typically processed at a centralized station for FR evaluation. The Ground-Station manages off-board processing using advanced deep learning (DL) models with the robots’ sensors. FRs at the station can then use an Augmented Reality (AR) interface, like smart glasses, to access near real-time disaster site information in a 3D reconstructed map. Key semantics may be annotated on the map, providing a safer and clearer method for operators to assess a crisis situation.

Integrating autonomous robots into FR tasks is challenging. Any relevant system, including the proposed robotic one, requires well-defined training activities, such as organized sessions, specialized workshops, and expert knowledge exchange, to advance this initiative. Additionally, a comprehensive examination of the technological, operational, organizational, and policy aspects of the current FR ecosystem is essential for effective training. In parallel, public awareness must be carefully assessed through educational programs and seminars to help citizens understand how to interact with AMRs during crises, emphasizing that robots are meant to assist, not to replace, human responders. Building a cooperative relationship between the general public and advanced technologies can foster trust, ultimately contributing to a more resilient society in disasters.

In this paper, an AMR-based system is introduced, termed `Autonomous Robotic Aid For Increasing
First Responders Efficiency' (TRIFFID\footnote{https://triffid-project.eu}). The system aims to establish an advanced technical framework by integrating cutting-edge AI tools and algorithms, potentially enabling robots to conduct remote reconnaissance and to provide critical information to FRs. Key challenges include developing autonomous navigation capabilities for independent and efficient robot movement, and implementing effective mission and path planning strategies to optimize task execution, while considering mission objectives and obstacles. Another significant challenge is the collection and interpretation of semantic data from diverse sensors, which requires processing and analysis to yield actionable insights. These challenges will be tackled using the latest technologies to improve robot capabilities and performance in various scenarios. Overall, this work aims to deliver a thorough examination of the challenges that the proposed system is crafted to address, as highlighted by its various use cases. Moreover, an in-depth analysis of the technical architecture and constituent elements of the system are presented. These findings may offer a significant illustration of how robotic platforms and artificial intelligence tools can be efficiently employed to assist FRs in their essential and often life-saving operations.

The paper is structured as follows: Section \ref{sec:related-work} reviews research and projects in disaster response robotics. Section \ref{sec:proposed-system} highlights the role of AMRs in emergency scenarios. Section \ref{sec:system-details} analyzes key system components, including navigation, communication, and perception modules. Section \ref{sec:use-cases} presents three practical real-world applications scenarios, where the system will be evaluated: wildfire response, urban flood management, and post-earthquake search and rescue. Finally, Section \ref{sec:conclusion} summarizes contributions and outlines future research directions.

\section{Related Work}
\label{sec:related-work}

\subsection{Research Studies}

Urban Search And Rescue (USAR) systems employ semi-autonomous control schemes, which incorporate hierarchical reinforcement learning and utilized sensors comparable to those in the proposed system. These systems also include manual and tele-operated fall-back mechanisms \cite{doroodgarLearningBasedSemiAutonomousController2014}.

The literature on the use of aerial AMRs focuses on an extensive examination of several topics, including platforms, sensors on board, Simultaneous Localization And Mapping (SLAM) methodologies, terrain coverage techniques, autonomous navigation strategies, and human-swarm interfaces \cite{recchiutoPostdisasterAssessmentUnmanned2018}. The detection of damaged infrastructure is also a subject of research, with studies investigating the use of UGV platforms to detect damage in simulated earthquake scenarios \cite{chenDetectionDamagedInfrastructure2019}.

In parallel, researchers investigate Multi-Robot Task Allocation (MRTA) problems in the context of flood response scenarios, assessing the performance of multiple UAV robots subject to range and payload constraints \cite{ghassemiMultirobotTaskAllocation2022}. The development of custom-made robots for earthquake scenarios is also an active area of research, with proposals for modular and snake-like robots that are lightweight and compact in size \cite{narayanSearchReconnaissanceRobot2022, jadejaSurvivorDetectionApproach2024}. Moreover, researchers have evaluated the deployment of Unmanned Surface Vehicles (USVs) alongside UAVs and UGVs, which require sophisticated frameworks to manage disaster scenarios \cite{pillaiHeterogeneousRobotsCollaboration2024}.

Indoor post-disaster human detection is crucial for FRs, and Micro-Aerial Vehicles (MAVs) possess the advantage of maneuvering in confined spaces. MAVs equipped with thermal cameras and real-time algorithms can precisely detect survivors, as well as map the scene for additional analysis \cite{tavasoliAutonomousPostdisasterIndoor2025}.

\subsection{International Projects}

\begin{table}
  \centering
  \caption{Key research and development efforts for deploying robotic solutions in post-disaster scenarios.}
  \begin{tblr}{colspec = {X[c, 1.5cm]|X[c]|X[c]|X[c]|X[c]|X[c]},
    row{1, 2} = {font=\bfseries},
        cell{1}{2} = {c=2}{c},
        cell{1}{4} = {c=3}{c},
        cell{1}{1} = {r=2}{c},
      }
    \toprule
    Work                                                      & Robots     &            & Setting    &            &              \\ \midrule
                                                              & UAV        & UGV        & Wildfire   & Flood      & Earthquake   \\ \midrule
    \cite{tavasoliAutonomousPostdisasterIndoor2025}           & \checkmark &            &            &            & \checkmark   \\
    \cite{pillaiHeterogeneousRobotsCollaboration2024}         & \checkmark & \checkmark &            & \checkmark &              \\
    \cite{jadejaSurvivorDetectionApproach2024}                &            & \checkmark &            &            & \checkmark   \\
    \cite{narayanSearchReconnaissanceRobot2022}               &            & \checkmark &            &            & \checkmark   \\
    \cite{chenDetectionDamagedInfrastructure2019}             &            & \checkmark &            &            & \checkmark   \\
    \cite{recchiutoPostdisasterAssessmentUnmanned2018}        & \checkmark &            &            &            & \checkmark   \\
    \cite{doroodgarLearningBasedSemiAutonomousController2014} &            & \checkmark &            &            & \checkmark   \\ \midrule
    CARMA \cite{CARMAProject2024}                             &            & \checkmark & \checkmark &            &              \\
    SILVANUS \cite{SILVANUSIntegratedTechnological2021}       & \checkmark & \checkmark & \checkmark &            &              \\
    CURSOR \cite{ristmaeCURSORSearchRescue2021}               & \checkmark & \checkmark &            &            & \checkmark   \\
    \midrule TRIFFID                                          & \checkmark & \checkmark & \checkmark & \checkmark & \checkmark & \\
    \bottomrule
  \end{tblr}
  \label{tab:system_comparison}
\end{table}

\begin{figure*}
  \centering
  \includegraphics[width=0.65\linewidth]{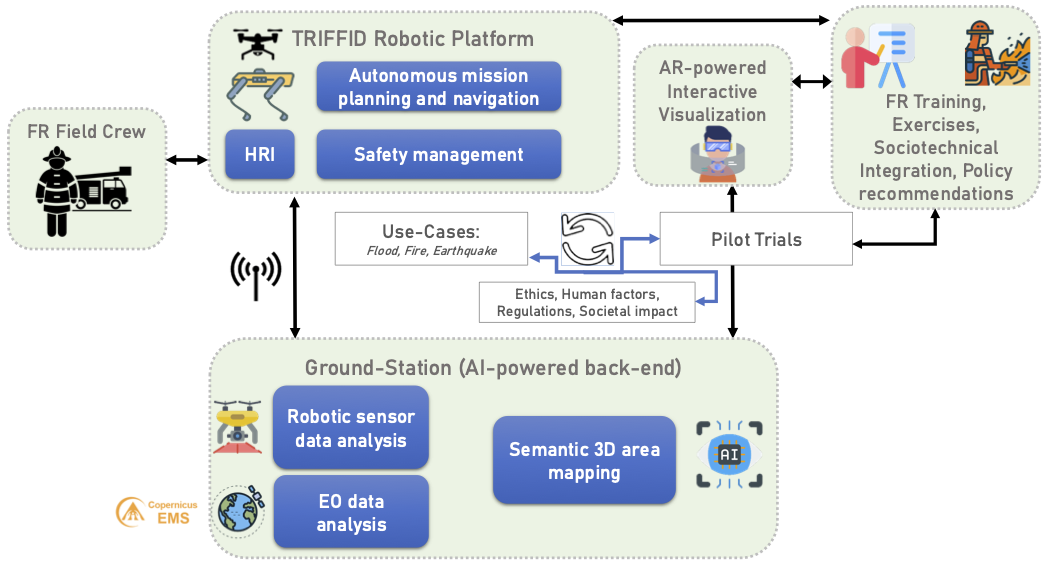}
  \caption{TRIFFID system functional architecture.}
  \label{fig:triffid-functional-arch}
\end{figure*}

Apart from dedicated and task-oriented research works, more large-scale and integrated robotic systems for disaster management have recently been launched.

SILVANUS is an ongoing project focused on developing an integrated wildfire management platform, using UAV and UGV robots, public Internet data, and IoT devices \cite{SILVANUSIntegratedTechnological2021}.

CARMA is an active project that uses off-road tracked and legged UGVs to assist FRs in their operations during disaster scenarios. These scenarios include urban zones affected by earthquakes, underground parking garage fires, cargo incidents, and Chemical, Biological, Radiological, Nuclear, and Explosive (CBRN-E) incidents \cite{CARMAProject2024}.

The CURSOR project develops a search and rescue kit that includes drones, minified robotic equipment, and advanced sensors. This kit is designed to reduce the time required for detection and rescue of victims trapped under debris, while enhancing the personal safety of the search and rescue team \cite{ristmaeCURSORSearchRescue2021}.

While the above projects have substantially contributed to the successful integration of robotics to FR procedures, the TRIFFID system proposes a harmonized solution of both a UAV and UGV for autonomous reconnaissance in three demanding real-world scenarios, namely wildfire, urban flood, and earthquake. Table \ref{tab:system_comparison} summarizes key research and development efforts for deploying robotic solutions in post-disaster scenarios.

\section{Proposed System}
\label{sec:proposed-system}

The 'Autonomous robotic aid for increasing first responders efficiency' system (entitled \textbf{TRIFFID}), whose overall functional architecture is illustrated in Fig. \ref{fig:triffid-functional-arch}, is composed of the following main components:

\begin{enumerate}
  \item A comprehensive robotic platform, composed of a hybrid autonomous UGV and a UAV robot,
  \item A centralized ground-station, which is equipped with an advanced AR interface to be interacted with by a human system operator,
  \item A custom wireless communication infrastructure, which will interconnect at low latency the robots, the FR field crew and the Ground-Station, and
  \item A smartphone app to be carried by each member of the human FR field crew
\end{enumerate}

The system is set up with a ground-station, a UAV launch pad, a UGV, and a remote system operator located near the local FR Base-of-Operations (BoO) at a mission's disaster site. The UGV accompanies the field crew within the disaster scene. The ground-station reports to the local Disaster Scene Commander (DCS), who directly supervises the BoO. Each field crew member carries a smartphone equipped with the system's application, offering different functionalities for: i) the field crew leader and ii) multiple field crew members.

The unified TRIFFID system aims to be seamlessly integrated into FR operational procedures. Its initial deployment begins when the FR team arrives at the disaster site and establishes a BoO. Immediately, local private wireless communications infrastructure is set up, deploying network hotspots in strategic locations to cover the disaster area wirelessly. This triggers the following four events:

\begin{enumerate}
  \item The operator interacting with the TRIFFID ground-station initiates the system, which receives and analyzes Earth Observation (EO) data in order to construct a preliminary, semantically annotated area map, if Internet connectivity is available at the BoO. This provides prior knowledge for understanding the damage and delimiting the concerned area. If no Internet connectivity is available, this preliminary map is manually provided by the responsible authorities.
  \item The FR field crew activates their TRIFFID smartphone app, which is updated with the preliminary area map as soon as it is ready.
  \item The ground-station operator sets any high-level mission details (e.g., flight path, including waypoints, terrain avoidance, no fly zones, etc.) to be issued to the robots.
  \item The UAV is deployed to aerially survey the disaster area and, thus, to update the semantic area map with real-time information.
\end{enumerate}

\begin{figure*}
  \centering
  \includegraphics[width=0.7\linewidth]{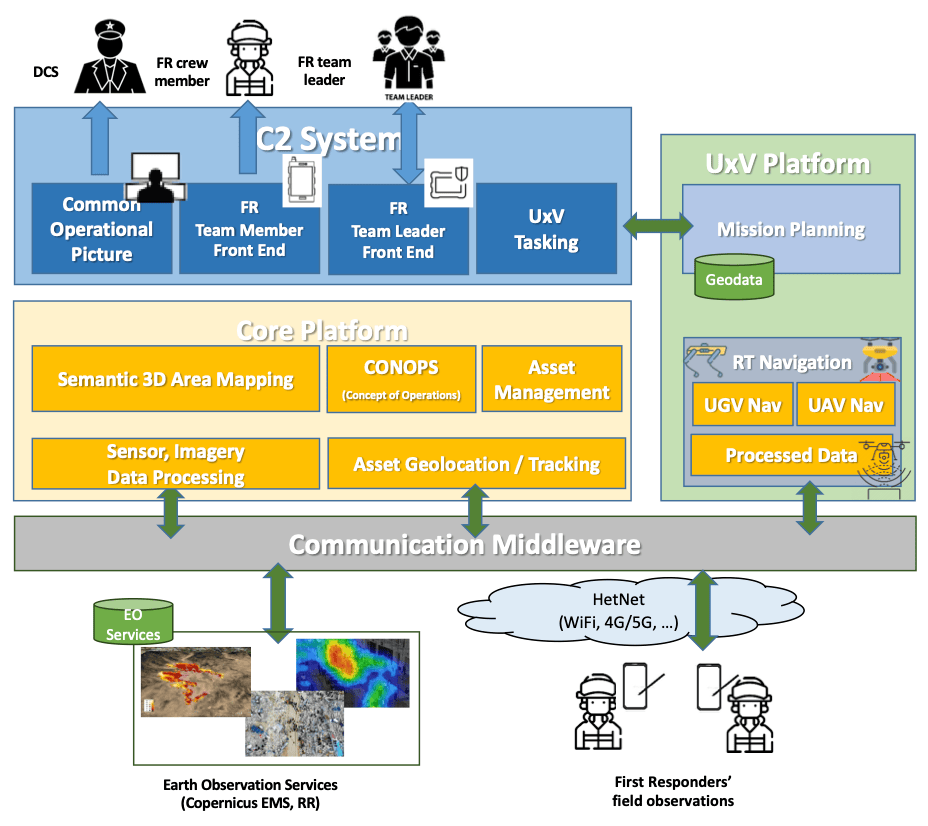}
  \caption{TRIFFID system technical architecture.}
  \label{fig:triffid-technical-approach}
\end{figure*}

At this stage, a preliminary map is developed for autonomous navigation, while the system operator plans UAV tasks through high-level commands at the ground-station. Simultaneously, the FR field crew prepares for deployment. After the UAV returns for battery recharging and the FR crew deploys to the disaster site with the autonomous UGV, the ground-station starts receiving data from the UGV’s sensors. A centralized 3D semantic area map is gradually created from this data, with updated versions periodically broadcasted to the TRIFFID smartphone app used by the FR field crew. The UGV typically follows the FR crew, but they can issue high-level commands from a predefined set, such as autonomously mapping, delivering supplies, searching for civilians, or checking for hazards. Human-Robot Interaction (HRI) is materialized via gestures or speech, with the smartphone app as a backup. The FR crew continues operations, while the UGV autonomously pursues its goal. If the UGV faces path challenges due to debris, it alerts the ground-station, prompting the remote operator to dispatch the UAV for a targeted aerial survey. The re-deployed UAV autonomously navigates to the specified area, surveys it, and sends data back to the ground-station, updating the map automatically. Eventually, the UGV receives the new map wirelessly and proceeds with its task if possible; otherwise, it executes a suitable fail-safe behavior (e.g., return to BoO, rejoin the FR crew, request new orders, etc.). This also occurs during communication failures. Under normal conditions, both the FR field crew (via the smartphone app) and the system operator (via the ground-station) can remotely monitor and influence the UGV. If desired, the system robots can switch from autonomous to teleoperated navigation. A backup human UAV pilot is available at the BoO for emergencies.

\section{Robot-powered Post-Disaster Reconnaissance}
\label{sec:system-details}

This section provides detailed information on the main components of the proposed system. To facilitate autonomous navigation and communication with the Base of Operations (BoO), advanced algorithms will be explored. The semantic perception of the deployed robots, as well as their interactions with FRs and victims, will be analyzed. The system’s overall technical architecture is depicted in Fig. \ref{fig:triffid-technical-approach}.


\subsection{Navigation and Communication}

Disaster scenarios, including wildfires, floods, and earthquakes, necessitate strong navigation-related modules. These include i) mission and task planning, ii) autonomous navigation techniques, iii) safety management to prevent harm and to ensure fault recovery, and iv) a low-latency, private, and heterogeneous wireless communications infrastructure for continuous interaction among the system’s components.

\subsubsection{\textbf{Mission and Task Planning}}

The initial module, 'Mission and task planning,' aims to autonomously organize the task plan for the system’s robots, based on disaster site information and high-level mission details provided by humans. These details may include goals, no-fly areas, waypoints, and terrain avoidance zones, and can be communicated via the BoO, direct HRI, or the FR smartphone app. The module will dynamically decompose goals into smaller tasks and will adapt the plan if necessary or if goals change. A three-layer structure will be supported: i) a top layer for mission planning, managing the current mission according to the FR strategy and specifications, ii) a global motion planning layer of each robot, integrating mission planner requests with robot capabilities to define exploration, search, and coverage behaviors in the assigned zone, and iii) a traversability and local navigation sub-module \cite{santamaria-navarroTerrainClassificationComplex2015} shall analyze terrain traversability and velocities. Pre-defined concepts of operations (CONOPS) for emergency response will optimally exploit the robotic platform and inserted into the ground station’s core one. UGV tasks will be managed through behavior trees \cite{iovinoSurveyBehaviorTrees2022} to enable on-board decisions based on the current scenario belief, while also considering external inputs from the FRs or the operator.

\subsubsection{\textbf{Autonomous Robot Navigation}}

The second module, 'Autonomous Robot Navigation,' will implement navigation algorithms for the system’s robots drawing inspiration from previous work \cite{santamaria-navarroDeploymentAutonomousLastMile2024, linardakisDistributedMazeExploration2024} but, this time, considering the dynamic nature of disaster sites. It will execute the task plan derived by the 'Mission and Task Planning' module at a low level. Methodologically, it will build on previous approaches \cite{aghaNeBulaTEAMCoSTARs2022, linardakisDistributedMazeExploration2024} to create a reliable, real-time navigation subsystem integrating perception, planning, control, and localization. This includes: i) one-shot harsh terrain traversability modeling using sensor data and local 3D mapping with the UGV’s LiDAR, extending the method from \cite{pinoProbabilisticGraphBasedRealTime2024} to account for noise, terrain elevation probability distribution, and neural network classification of similar ground types. ii) Dynamic path planning and obstacle avoidance by adapting a sliding-window non-linear Model Predictive Control (MPC) method for trajectory prediction, incorporating obstacles as constraints \cite{lopezDynamicTubeMPC2019}, to ensure safe UVG navigation in harsh environments. iii) Higher-level robot behavior structuring using behavior trees \cite{macenskiMarathon2Navigation2020} to integrate navigation with high-level task management, such as HRI tasks. The current scene representation will be obtained from the 'Non-visual Data Analysis and Information Fusion' module. The UGV robot will be the most challenging implementation, as navigation is more straightforward for the UAV; thus, off-the-shelf components shall be adopted and customized for the UAV. Both UGV and UAV methods will be tuned for harsh conditions (e.g., smoke, dust, humidity, high temperature, etc.).

\subsubsection{\textbf{Safety Monitoring and Fault Recovery}}

The 'Safety monitoring and fault recovery' module evaluates if a UGV or UAV can achieve a goal or should be aborted and returned to its base. It will include fault diagnostics and failsafe routines, extending previous methodologies \cite{santamaria-navarroResilientAutonomousNavigation2022} to address redundancy and heterogeneity in sensing and task execution. The module will anticipate and detect failures, adapting behavior for safety. This will involve: i) monitoring robot health via hardware drivers, ii) detecting failures by comparing with nominal data, triggering alarms based on severity, and iii) recovering modules through resiliency logic. Redundancy will be used for switching problematic modules to functional ones and for reinitializing the failed system. Component status will be communicated to the 'Mission and task planning' and 'Autonomous Robot Navigation' modules for mission adjustments. Decisions and health reports will be sent to the ground station for remote checks. In case of serious errors with communication loss, locally executed behavioral fail-safes will be activated. In this way, FRs and the operator are notified of errors to switch to manual teleoperation, if desired.

\subsubsection{\textbf{Communications}}

The fourth module, 'Communications', will create a private low-latency wireless network. A heterogeneous architecture will integrate 4G, Wi-Fi, and possibly 5G, using existing disaster site infrastructure for redundancy. Network hotspots will be strategically deployed as needed. Security will be prioritized by design, and low latency will be improved with dynamic load balancing algorithms \cite{khayatWeightedClusterSUAV2023}, prioritizing real-time video over non-critical data. The network will switch to alternative options if a network type becomes unavailable, with real-time management and monitoring tools from the ground-station. Scalability for new technologies will be a key consideration, optimizing communications efficiency through diverse connectivity options.

\subsection{Perception and Human-Robot Interaction}

This component aims to improve real-time deep neural networks (DNNs), knowledge graphs (KG), and augmented reality (AR) techniques for robot perception \cite{rodisMultimodalExplainableArtificial2024}. These enhancements will facilitate HRI, human-computer interaction (HCI), and centralized information fusion. Consequently, both robots and the ground station operator will achieve a semantically rich and precise understanding of the disaster site. The following modules will be developed: i) visual semantic environment mapping using sensors and Earth Observation (EO) analysis, ii) non-visual sensor data analysis, denoising, geolocalization, and multimodal information fusion at the ground station, iii) verbal communication and gesture recognition for HRI in the field with FRs and civilians, suitable for harsh conditions, and iv) advanced AR-based interactive visualization for the ground station operator.

\subsubsection{\textbf{Visual Scene Analysis and EO Data Analysis}}

This module, 'Visual scene analysis and EO data analysis,' will cover techniques for 3D visual SLAM using LiDAR data and semantic 3D SLAM from RGB camera inputs \cite{khayatWeightedClusterSUAV2023}. Models for entities will facilitate the SLAM subsystem to reconstruct the scene \cite{liuRegularizedDeepSigned2022}, supporting the mission planning module. Generative approaches, like diffusion models \cite{alimisisAdvancesDiffusionModels2025}, will be assessed for completing missing areas within the derived 3D point cloud under various conditions, occlusions, and limited visibility. Scene semantics will be extracted by specialized object detection and recognition DNNs from RGB and thermal inputs, relevant to use cases, such as trees, water, fire, smoke, vehicles, debris, victims, FRs, buildings, and powerlines. For 3D SLAM from LiDAR data, existing estimation frameworks, e.g., WOLF \cite{solaWOLFModularEstimation2022}, based on factor graphs, will be extended to aid 3D LiDAR scan fusion. In parallel, this module will also concentrate on creating DNNs for semantic segmentation of wildfire burnt zones, smoke clouds, flooded areas, and earthquake-damaged city regions in EO data. The models will feature novel architectural changes and training objectives that will adopt various learning paradigms, including supervised, unsupervised, adversarial, and reinforcement learning. This strategy seeks to extract more information from a fixed-size dataset, by either improving test-time accuracy or achieving the same accuracy with a smaller, more efficient model. Moreover, in order to tackle the scarcity of large-scale datasets and the evolving nature of post-disaster sites, data augmentation, self-supervised visual learning \cite{konstantakosSelfsupervisedVisualLearning2025}, out-of-distribution detection, and domain generalization techniques will be applied.

\subsubsection{\textbf{Non-visual Data Analysis and Information Fusion}}

The second module, 'Non-visual data analysis and information fusion,' will generate, update, and maintain a merged, multiview, semantically annotated 3D area map. It will use Kalman filtering to integrate LiDAR- and RGB-based maps from each robot, considering non-simultaneous deployment. The final map will be annotated with unused semantic outputs and will include geolocalized non-visual measurements from robot sensors and FR smartphones, updated in near-real-time. Additionally, this module will advance DNNs for real-time environmental sound analysis and processing, adapting speech recognition for disaster site conditions. Audio cues will aid in semantic annotation of the unified 3D area map, with speech recognition capabilities employed in the third module.

\subsubsection{\textbf{HRI and Human Behavior Analysis}}

The third module, 'HRI and human behavior analysis,' will utilize real-time DNNs for semantic recognition of FR and civilian emotions, activities, and gestures \cite{adaloglouComprehensiveStudyDeep2022} under disaster site conditions, like occlusions and low visibility. Multimodal generative learning strategies will address these issues, focusing on distress signs. Nearby human motion and intention will be predicted for long-term horizons \cite{tankeIntentionbasedLongTermHuman2021} using a stratified Transformer DNN for efficient spatiotemporal analysis \cite{liangDualFormerLocalGlobalStratified2022}. Sophisticated training objectives and architectural modifications (Transformer, LSTM, CNN) will be explored to enhance accuracy \cite{doshiMultiTaskLearningVideo2022,papadopoulos2022user}. Outputs will integrate with the current semantic 3D area map to populate a Knowledge Graph (KG) using a novel ontology, facilitating real-time KG reasoning for contextualized robot behavior near humans \cite{papadopoulosOpenExpandableCognitive2021}. To achieve this, the IEEE RAS ontology \cite{olszewskaOntologyAutonomousRobotics2017} shall be extended by formalizing representations of knowledge about health state and needs of target humans, physiological state or current operational status of the FRs. The KG engine will adapt robot responses to HRI inputs via visual gesture recognition \cite{linardakisSurveyHandGesture2025} and LLM-enabled verbal communication. A pretrained LLM will be fine-tuned for disaster site conditions \cite{zhangLargeLanguageModels2023}, allowing natural-language instructions to the UGV by FR crew members \cite{singhProgPromptGeneratingSituated2023} and verbal assessments of stranded civilians’ needs. Moreover, the KG will selectively enrich/update the 3D map online and vice versa, extending previous work \cite{umbricoEnhancedCognitionAdaptive2022}.

\subsubsection{\textbf{Intuitive AR-based Human-Computer Interfaces}}

The final module, 'Intuitive AR-based human-computer interfaces,' will provide the fully annotated 3D semantic area map from the 'Non-visual data analysis and information fusion' module to the ground-station operator. This will be performed through a real-time interactive visualization tool offering multiple views, accessible via an advanced mobile AR interface and a backup desktop Graphical User Interface (GUI). Both will maintain connections to updated maps, with the AR interface optimizing situational awareness and cognitive load management for efficiency. Super-resolution approaches in the AR system will enhance visual fidelity \cite{liSuperResolutionAugmented2022}.

\section{Use-Cases}
\label{sec:use-cases}

\begin{figure*}[t]
  \centering
  \begin{tabular}{ccc}
    \begin{tabular}{c}
      \begin{subfigure}[t]{0.25\textwidth}
        \includegraphics[width=\textwidth]{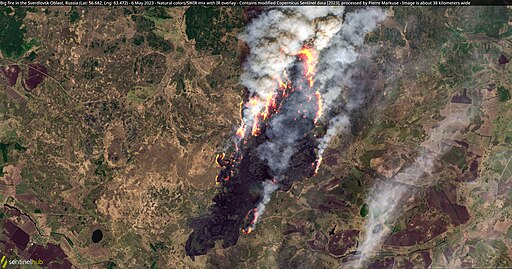}
        \caption{}
        \label{fig:uc-wildfire}
      \end{subfigure}
    \end{tabular} &
    \begin{tabular}{c}
      \begin{subfigure}[t]{0.25\textwidth}
        \includegraphics[width=\textwidth]{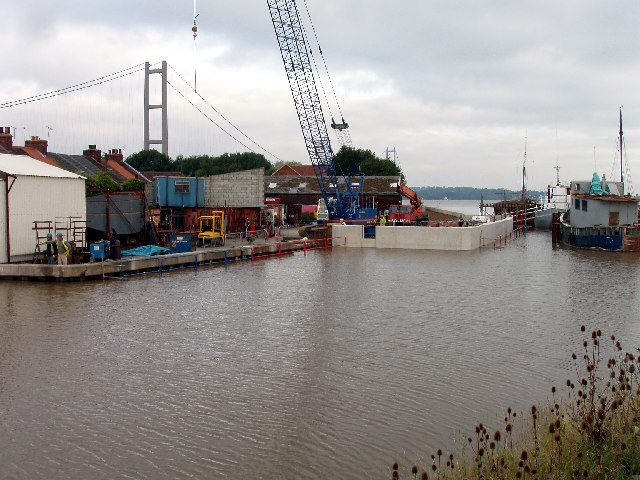}
        \caption{}
        \label{fig:uc-urban_flood}
      \end{subfigure}
    \end{tabular} &
    \begin{tabular}{c}
      \begin{subfigure}[t]{0.25\textwidth}
        \includegraphics[width=\textwidth]{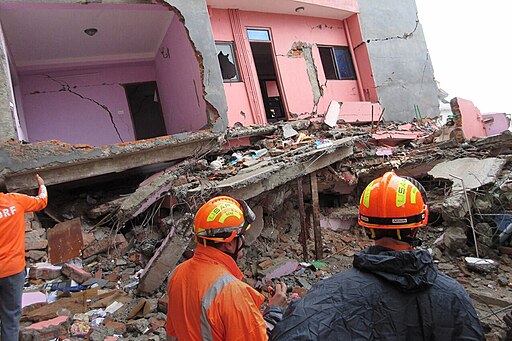}
        \caption{}
        \label{fig:uc-usar}
      \end{subfigure}
    \end{tabular}
  \end{tabular}
  \caption{Indicative visual content from TRIFFID's supported use cases: i) Wildfire in suburban environments \cite{markuseContainsModifiedCopernicus2023}, ii) Urban flood \cite{wrightRaisingFloodDefences2005}, and iii) USAR after earthquake \cite{defensieNederlandsHetWas2015}.}
  \label{fig:ucs}
\end{figure*}

Recent research \cite{elkadyWhatEmergencyServices2022} shows that FRs need credible disaster scene information quickly. Common natural disaster risks include floods, extreme weather, wildfires, epidemics, and earthquakes \cite{directorate-generalforeuropeancivilprotectionandhumanitarianaidoperationsechoeuropeancommissionOverviewNaturalManmade2021}. Given urbanization risks, the system will be demonstrated through three use-cases for European urban populations. In this context, TRIFFID is using a robotic platform as a force multiplier to support FRs, improve situational awareness, and enhance operational efficiency. This approach reduces risks by minimizing human exposure and optimizing resource allocation. Fig. \ref{fig:ucs} shows representative images for each use case that the TRIFFID system will be designed to address.

\subsection{Wildfire in Suburban Environments}

A forest fire is nearing a highly explosive industrial facility with three active fronts and a path towards the plant. The situation is worsened by smoke and winds over 70 km/h. Firefighters need a near-real-time semantic map from the proposed system, which highlights access paths, roads, temperature, plant dehydration, and wind speed and direction. Reconnaissance occurs in two stages: a UAV collects aerial information, and a UGV, part of the FR field crew, operates autonomously for collecting detailed data. The TRIFFID mission must be executed quickly before the fire reaches the facility. The semantic map, generated in near-real-time, is accessible at the BoO for the remote operator and through a smartphone app for the FR crew.

\subsection{Urban Flood}

An ongoing dual challenge of urban flooding and a crisis at a chemical plant, worsened by flood-compromised cooling systems, is occurring. There is a risk of industrial spills of hazardous gases. Initially, the UAV surveys flooded areas, identifying hazards, like submerged vehicles, collapsed structures, and downed power lines. It also detects stranded citizens and highlights the urgent situation at the chemical plant, by identifying cooling water loss. The BoO relays relevant near real-time updates and instructions to the FRs via their smartphone app. In the second stage, the UGV, with a secondary FR field crew, safely delivers supplies to the primary FR crew, including an emergency cooling system component, using the aerial overview. The primary FRs then navigate within the flooded areas to reach both the affected community and the chemical plant. They periodically send the UGV for autonomous advanced reconnaissance to detect potential spilled hazardous gases and stranded citizens. As they proceed, the FRs rescue victims and restore the cooling systems using the UGV’s supplies. Throughout this stage, the UAV is occasionally redeployed in a targeted manner to provide updates on emerging hazards.

\subsection{USAR after Earthquake}

A team of FRs is dispatched to conduct an urban search and rescue operation in the chaotic and hazardous aftermath of an earthquake. The FR field crew deploys the UGV to semantically map a critical city area by navigating through debris and selecting the safest and most efficient route, while avoiding obstacles. Concurrently, the UAV performs autonomous aerial surveillance to identify survivors and to assess damages, facilitating and enabling mapping requests for specific areas. Critical objects, such as injured individuals, fire, water sources, and obstacles requiring removal, are automatically identified. The semantic map, dynamically created in near real-time, is accessible at the BoO for the operator’s interaction, but it is also available to the FR crew through their smartphone application.

\section{Conclusion}
\label{sec:conclusion}

This paper presents a comprehensive system for integrating autonomous mobile robots into FR operations, focusing on enhancing reconnaissance during disasters. The TRIFFID system combines unmanned aerial and ground vehicles with advanced AI tools to create a unified technical framework that significantly boosts situational awareness and operational efficiency. By integrating state-of-the-art technologies in navigation, perception, and human-robot interaction, the system enables autonomous reconnaissance, while maintaining human oversight. The supported use cases of wildfires, urban floods, and post-earthquake scenarios demonstrate the system’s adaptability and potential impact across various disasters. TRIFFID will provide FRs with real-time semantic mapping, autonomous navigation, and augmented reality interfaces, addressing operational needs and reducing personnel exposure to hazards.

Regarding future steps, successful implementation of this system could mark a significant advancement in disaster response capabilities. The modular architecture ensures adaptability to different scenarios, while the focus on human-robot interaction and safety monitoring guarantees reliability for real-world deployment. As climate change and urbanization increase the frequency and severity of natural disasters, systems like TRIFFID will become increasingly valuable. Future research should focus on extensive field testing, refinement of human-robot interaction interfaces, and development of additional use cases. The integration of robotic platforms into standard emergency response protocols signifies a promising direction for enhancing the effectiveness and safety of FR operations in challenging disaster scenarios.

\printbibliography

\end{document}